\pdfoutput=1

\documentclass[11pt]{article}

\usepackage{acl}

\usepackage{times}
\usepackage{latexsym}
\usepackage{enumitem}

\usepackage[T1]{fontenc}

\usepackage[utf8]{inputenc}

\usepackage{microtype}
\usepackage{graphicx}
\usepackage{enumitem}

\usepackage{multirow}
\usepackage{amssymb}

\usepackage{multirow}
\usepackage{footmisc}
\usepackage{subcaption,siunitx,booktabs}
\usepackage{caption}
\usepackage{bbm}
\usepackage{todonotes}
\usepackage{amsmath}

\usepackage{balance}
\usepackage{enumitem}
\usepackage[ruled]{algorithm2e}
\SetKwComment{Comment}{$\triangleright$\ }{}

\SetKwInput{KwInput}{Input}                
\SetKwInput{KwOutput}{Output}              

\renewcommand{\todo}[1]{\iffalse #1 \fi[TODO]}

\newcommand{\ie}{\textit{i}.\textit{e}.}

\newcommand{\vpara}[1]{\vspace{0.05in}\noindent\textbf{#1 }}

\makeatletter
\def\thanks#1{\protected@xdef\@thanks{\@thanks\protect\footnotetext{#1}}}
\makeatother

\title{End-to-End Modeling via Information Tree for One-Shot Natural Language Spatial Video Grounding}

\author{Mengze Li$^1$, Tianbao Wang$^1$, Haoyu Zhang$^1$, Shengyu Zhang$^1$, Zhou Zhao$^{2,3*}$, \\ {\bf Jiaxu Miao$^{1*}$, Wenqiao Zhang$^{1*}$, Wenming Tan$^4$, Jin Wang$^4$, Peng Wang$^5$,} \\ {\bf Shiliang Pu$^4$, Fei Wu$^{2,3*}$ } \thanks{First author.} \thanks{mengzeli@zju.edu.cn} \thanks{} \thanks{$^*$Corresponding author.} \thanks{zhaozhou@zju.edu.cn} \thanks{jiaxu.miao@yahoo} \thanks{wenqiaozhang@zju.edu.cn} \thanks{wufei@zju.edu.cn} \\
$^1$Zhejiang University
$^2$Shanghai Institute for Advanced Study of Zhejiang University \\ $^3$Shanghai AI Laboratory $^4$Hikvision Research Institute \\ $^5$Northwestern Polytechnical University}

\begin{document}
\maketitle
\begin{abstract}
  Natural language spatial video grounding aims to detect the relevant objects in video frames with descriptive sentences as the query. In spite of the great advances, most existing methods rely on  dense video frame annotations, which require a tremendous amount of human effort.
  To achieve effective grounding under a limited annotation budget, we investigate one-shot video grounding, and learn to ground natural language in all video frames with solely one frame labeled, in an end-to-end manner.
  One major challenge of end-to-end one-shot video grounding is the existence of videos frames that are either irrelevant to the language query or the labeled frames. 
  Another challenge relates to the limited supervision, which might result in ineffective representation learning. To address these challenges, we designed an end-to-end model via \underline{I}nformation \underline{T}ree for \underline{O}ne-\underline{S}hot video grounding (IT-OS). Its key module, the information tree, can eliminate the interference of irrelevant frames based on branch search and branch cropping techniques. In addition, several self-supervised tasks are proposed based on the information tree to improve the representation learning under insufficient labeling. Experiments on the benchmark dataset demonstrate the effectiveness of our model.
\end{abstract}

\section{Introduction}

\begin{figure}[t]
  \centering
   \includegraphics[width=1.0\linewidth]{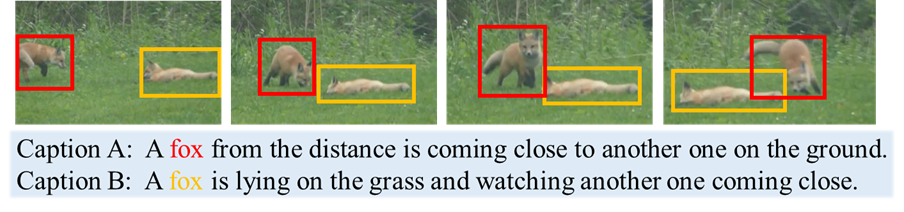}

   \caption{An example of spatially grounding natural language in video frames.}
   \label{fig:1_1}
\end{figure}

Natural language spatial video grounding is a vital task for video-text understanding \cite{luo2017comprehension, zhou2019grounded, hu2019you, Zhang_Tan_Yu_Zhao_Kuang_Liu_Zhou_Yang_Wu_2020, li2021adaptive}, which aims to detect the objects described by the natural language query from each video frame, as shown in Figure~\ref{fig:1_1}. 
There is a substantial and rapidly-growing research literature studying this problem with dense annotations \cite{li2017,yamaguchi2017spatio,sadhu2020video}, where each frame that contains objects relevant to the language query will be manually labeled with bounding boxes. Obviously, such annotations require tremendous human effort and can hardly be satisfied in real-world scenarios. Recently, some works have investigated weakly-supervised video grounding with solely the video-text correspondence rather than object-text annotations \cite{huang2018finding,chen2019object,shi2019not,chen2019weakly,zhou2018weakly}. However, the performance is less satisfied with such weak supervision. In practice, we are more likely to have a limited annotation budget rather than full annotation or no annotation. In addition, as humans, after experiencing the language query and one frame object paired together for the first time, we have the ability to generalize this finding and identify objects from more frames. Towards this end, we investigate another practical problem setting, \ie, one-shot spatial video grounding, with solely one relevant frame in the video labeled with bounding boxes per video.

Existing methods that are devised for supervised video grounding are not directly applicable to this novel setting. We summarize several critical challenges:
\begin{itemize}[leftmargin=*]
	\item  On the one hand, most of them incorporate a multi-stage training process, \ie, firstly training a clip localization module, and training an object localization module in the second stage. However, in one-shot spatial video grounding, there are no temporal annotations, which indicate the start/end time of the relevant clip, to train the clip localization module. Moreover, many of them extract video features in a pre-processed manner using feature extractor or object detector pretrained on large-scale datasets. However, independent modeling limits the cooperation of different modules, especially when the labels are few. Therefore, it is in urgent need to derive an end-to-end training framework for one-shot spatial video grounding.
	\item  On the other hand, there are video frames that are either irrelevant to the natural language query or the labeled frames. These irrelevant frames might increase the computation complexity of end-to-end training, and bring confounding between the frame label and (irrelevant) visual features. 
	\item Lastly, with fewer supervision signals, deep representation learning might become error-prone or easily under-fitting, especially for end-to-end training.
\end{itemize}
  



To address these challenges, we devise an end-to-end model via the \underline{I}nformation \underline{T}ree for the \underline{O}ne \underline{S}hot natural language spatial video grounding (IT-OS). Different from previous works, we design a novel tree structure to shield off the one-shot learning from frames that are irrelevant to either the language query or the labeled frame. 
We devise several self-supervised tasks based on the tree structure to strengthen the representation learning under limited supervision signals. Specifically, the calculation processes of the key module, information tree, contains four steps: 
(1) To construct the information tree, we view video frame features as nodes, and then compress the adjacent nodes to non-leaf nodes based on the visual similarity of themselves and the semantic similarity with the language query;
(2) We search the information tree and select branch paths that are consistently relevant to the language query both in the abstractive non-leaf node level and in the fine-grained leaf node level; 
(3) We drop I) the leaf nodes that do not belong the same semantic unit with the labeled node; and II) the non-leaf nodes on the low relevance branch paths. We also down-weight the importance of the leaf nodes that belong to the same semantic unit with the labeled node but are on the low relevance paths;
(4) Finally, we input the extracted and weighted information to the transformer, and conduct training with the one-shot label and self-supervised tasks, including masked feature prediction and video-text matching. We note that both the information tree and the transformer are jointly trained in an end-to-end manner.


We conduct experiments on two benchmark datasets, which demonstrate the effectiveness of IT-OS over state-of-the-arts. Extensive analysis including ablation studies and case studies jointly demonstrate the merits of IT-OS on one-shot video grounding.
Our contributions can be summarized as follows:
\begin{itemize}[leftmargin=*]
  \item To the best of our knowledge, we take the initiative to investigate one-shot natural language spatial video grounding. We design an end-to-end model named IT-OS via information tree to address the challenges brought by limited labels.
  \item By leveraging the language query, several novel modules on the information tree, such as tree construction, branch search, and branch cropping, are proposed. Moreover, to strengthen the deep representation learning under limited supervision signals, we introduce several self-supervised tasks based on the information tree.
  \item We experiment with our IT-OS model on two benchmark datasets. Comparisons with the state-of-the-art and extensive model analysis jointly demonstrate the effectiveness of IT-OS.
\end{itemize}

\section{Related works}

\vpara{Natural Language Video Grounding.}
Among numerous multimedia understanding applications~\cite{Zhang_Jiang_Wang_Kuang_Zhao_Zhu_Yu_Yang_Wu_2020,Zhang_Tan_Zhao_Yu_Kuang_Jiang_Zhou_Yang_Wu_2020,
zhang2021consensus,zhang2021magic,
zhang2020relational,kai2021learning, zhang2020counterfactual}, natural language video grounding has attracted the attention of more and more researchers recently. There are mainly three branches, temporal grounding[\cite{ross2018grounding,lu2019debug,zhang2019cross,lin2020weakly,lin2020moment,zhang2021parallel,li2022compositional,gao2021relation, yang2021deconfounded}], spatio-temporal grounding[\cite{tang2021human,zhang2020object,zhang2020does,su2021stvgbert}], and spatial grounding. We focus on the last one.

Deep neural network has convincingly demonstrated high capability in many domains \cite{wu2020biased, wu2022learning, guo2021semi, li2020multi, li2020ib, li2020unsupervised},
especially for video related tasks \cite{miao2021vspw, miao2020memory, xiao2020visual, xiao2021video}, like video grounding.
For example,\cite{li2017} use the neural network to detect language query related objects in the first frame
and track the detected object in the whole video. Compared to it, \cite{yamaguchi2017spatio} and \cite{vasudevan2018object} go further. 
They extract all the object proposals through the pretrained detector, and choose the right proposal described in the text.

Supervised training for the natural language video object detection needs high labeling costs. To reduce it, some researchers pay attention to weakly-supervised learning fashion using multiple instances learning(MIL) method \cite{huang2018finding,chen2019object,shi2019not,chen2019weakly,zhou2018weakly,wang2021weakly}transfers contextualized knowledge in cross-modal alignment to release the unstable training problem in MIL. Based on contrastive learning \cite{zhang2021reconstrast}, \cite{da2021asynce} proposes an AsyNCE loss to disentangle false-positive frames in MIL, which allows for mitigating the uncertainty of from negative instance-sentence pairs. Weakly supervised false-positive identification based on contrastive learning has witnessed success in some other domains~\cite{Zhang_Yao_Zhao_Chua_Wu_2021,yao2021contrastive}

\vpara{One-shot Learning for Videos.} One-shot learning has been applied in some other video tasks.
\cite{yang2018one} proposes a meta-learning-based approach to perform one-shot action localization by capturing task-specific prior knowledge.
\cite{wu2018exploit} investigates the one-shot video person re-identification task by progressively improving the discriminative capability of CNN via stepwise learning.
Different from these works, \cite{caelles2017one} and \cite{meinhardt2020make} define the one-shot learning as only one frame being labeled per video.
 Specifically,
\cite{caelles2017one} use a fully convolutional neural network architecture to solve the one-shot video segmentation task.
\cite{meinhardt2020make} decouple the detection task, and uses the modified Mask-RCNN to predict local segmentation masks.
Following this setting, we investigate one-shot natural language spatial video grounding, and devise a novel information-tree based end-to-end framework for the task.

\section{Method}


\subsection{Model Overview}

\vpara{Problem Formulation.} Given a video $V=\{v^i\}_{i=1,2,\dots,I}$ and a natural language query $C$, spatial video grounding aims to localize the query-described object from all the objects $O^i=\{o_j^i\}_{j=1,2,\dots,J}$ for each frame. 
$I$ denotes the frame number of the video, and the $J$ is the object number in the video. 
In \textit{\textbf{one-shot}} spatial video grounding, 
solely one frame $v^i$ in video $V$ is labeled with the region boxes of the target objects $O^i$. 

\begin{figure*}[t] \begin{center}
  \includegraphics[width=0.9\textwidth]{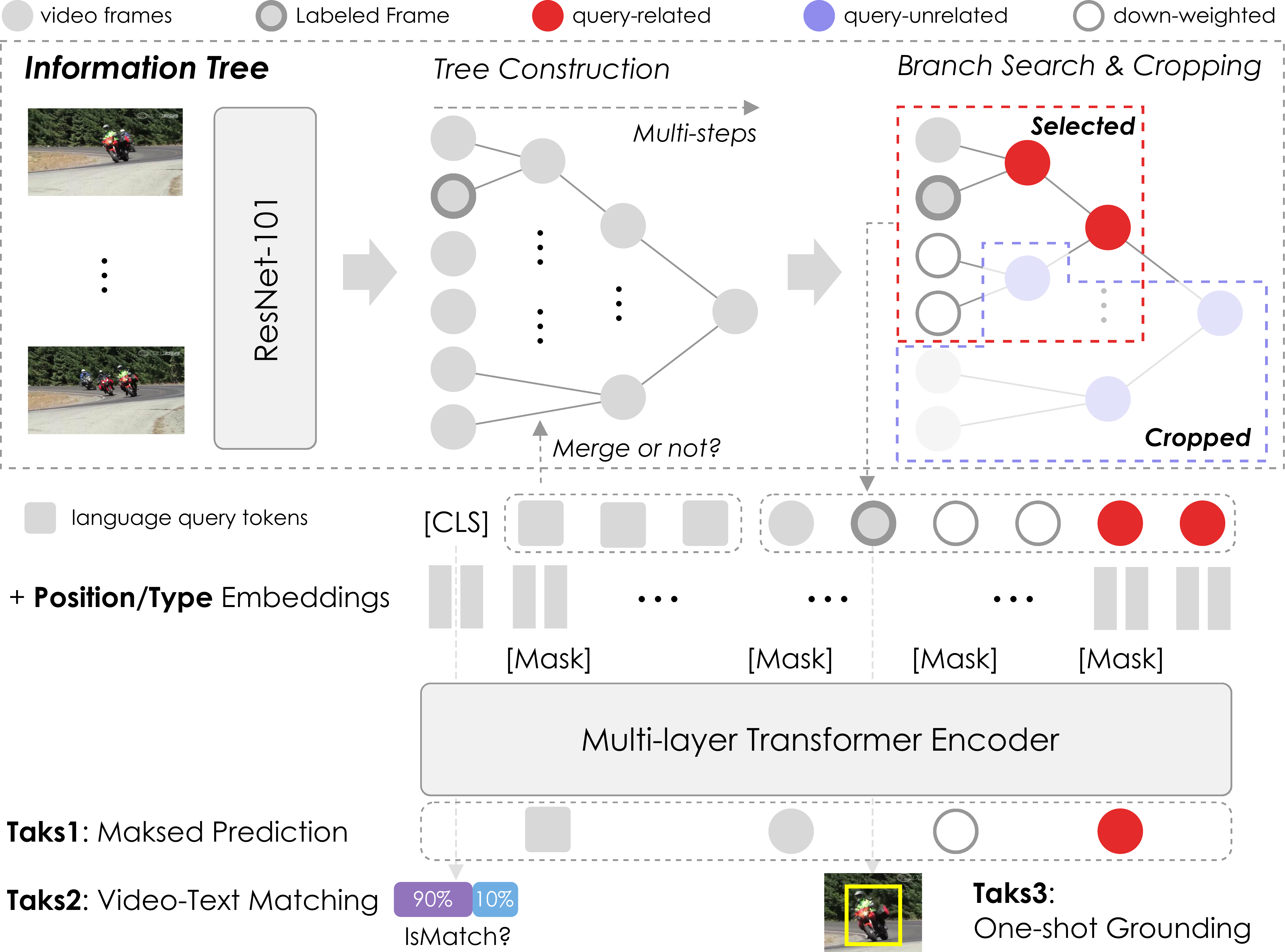}
  \caption{
  The overall schema of the proposed end-to-end one-shot video grounding via information tree (IT-OS), which contains query-guided tree construction, query-based branch search \& cropping, and a transformer encoder enhanced by self-supervised tasks.
}
\label{fig:schema}
\end{center} \end{figure*} 

%

\vpara{Pipeline of IT-OS.} As shown in Figure~\ref{fig:schema}, there are mainly four steps involved in the end-to-end modeling of IT-OS:

\begin{itemize}[leftmargin=*]
\item Firstly, we extract the features from the input video and the input caption. Specifically, for the video, we use ResNet-101\cite{he2016deep} as the image encoder to extract the frame feature maps; for the language query, we employ a language model Roberta\cite{liu2019roberta}. Both the vision encoder and the language encoder are jointly optimized with the whole network.
\item Secondly, we build the information tree to get the representation of the video. The information tree is built upon the frame feature maps, which are the leaf nodes. Leaf nodes will be further merged based on the relevance between node-node and node-query to have non-leaf and root nodes. 
Nodes on unnecessary branches will be deleted conditioned on the language query. 
\item Thirdly, we utilize the transformer encoder to reason on the remaining nodes and language features. Upon the transformer, we devise two self-supervised tasks, \ie, masked feature modeling, and video-text matching, which enhances the representation learning under limited labels.
\end{itemize}

\vpara{Prediction and Training.} We follow the common prediction and training protocol of visual transformers used in other object detection models \cite{wang2021end}.
We input the embedding parameters $E_{de}$ and the multi-model features $F_{de}$ generated by the transformer encoder into the transformer decoder $D$. Then, the decoder $D$ outputs possible prediction region features for each frame. For each possible region, a possibility $P$ and a bounding box $B$ are generated.
\begin{equation}
  P, B = D(F_{de}, E_{de}),
  \label{eq:decoder_pred}
\end{equation} 
We choose the box $B$ with the highest possibility value $P$ for each frame as the target box. 

During the training process, we first calculate the possible prediction regions. Then, we match the possible regions with the target boxes, and choose the best match for each frame. Finally, use the match to train our IT-OS model.

\subsection{Information Tree Module}
In this section, we will elaborate the information tree modules in detail. We will illustrate how to construct the information tree, how to extract critical information from it and how to design the self-supervised learning based on the tree. To ease the illustration, we take the $6$ frames as an example, and show the process in Figure~\ref{fig:schema}.

\subsubsection{Tree Construction} \label{sec:treeconstruct}

%

Given the frame features generated by the CNN, we build the information tree by merging adjacent frame features in the specified order. 
Specifically, the frame features output by the image encoder are the leaf nodes $N=\{n^i\}_{i=1}^{2M}$. A sliding window of size 2 and step 2 is applied on these nodes and nodes in the window are evaluated to be merged or not.


We calculate the \textit{semantic relevance difference} between each node pair with the language query, and get the \textit{visual relevance} between the nodes in each pair. 
For the visual relevance calculation, we max-pool the feature maps of the $i$ node pair to have the feature vector $f_v^{2i-1}$ and $f_v^{2i}$.
And then, we compute the cosine similarity $r_{vv}^i$ between $f_v^{2i-1}$ and $f_v^{2i}$ to be the visual relevance.
Next, we calculate the semantic relevance $r_{tv}^{2i-1}$ and $r_{tv}^{2i}$ between the text feature $f_t$ and the nodes of $i$ node pair:
\begin{equation}
  r_{tv}^{2i-1}=\sigma((w_{t}*f_t)*(w_{v}*f_v^{2i-1})^T),
  \label{eq:text_node_attention_1}
\end{equation} 
\begin{equation}
  r_{tv}^{2i}=\sigma((w_{t}*f_t)*(w_{v}*f_v^{2i})^T),
  \label{eq:text_node_attention_2}
\end{equation} 
where the $w_t$ and $w_v$ are learnable parameters, and $\sigma$ is the sigmoid activation function.

The semantic relevance difference $d_{tv}^i$ between the $i$th paired nodes is:
\begin{equation}
  d_{tv}^i=|r_{tv}^{2i-1}-r_{tv}^{2i}|+\gamma*r_{vv}^i,
  \label{eq:video_node_attention}
\end{equation} 
where the $\gamma$ is the hyperparameter.


With the relevant difference value, we rank the node pairs and pick out the top $\lambda$. The $\lambda$ is a hyperparameter, which can be set as a constant or a percentage. We merge the node pairs:
\begin{equation}
  n^{new}=w_{mg}*(n^{2i-1}+n^{2i})+b_{mg},
  \label{eq:node_merge}
\end{equation} 
where the $w_{mg}$ and $b_{mg}$ are trainable.
Finally, The new node $n^{new}$ replace the old nodes $n^{2i-1}$ and $n^{2i}$ in the queue. Repeat the process until there is only one node in the queue. Saving all nodes in the process and the composite relationship between nodes generated in the merging process, we get the information tree. 


\subsubsection{Branch Search}
%


We use a branch to denote a subtree. To filter critical local and global information, we perform branch search and selection. We firstly select branches that contain leaf nodes less than $\delta_{max}$ and more than $\delta_{min}$. $\delta_{max}$ and $\delta_{min}$ are hyperparameters. We calculate the \textit{semantic relevance} of branches' root nodes and the language query based on Equation \ref{eq:text_node_attention_1}.

\vpara{Training.} During training, we directly select the branch that contains the labeled leaf node and the root node with the highest semantic relevance. This selection improves the training efficiency.

\vpara{Inference.} During inference, all frames should be processed. We conduct an iterative search with multiple search steps. For each step, we select the branch with the highest semantic relevance and remove the selected branch from the information tree. After the search, we have multiple selected branches and each branch will be forwarded to the following processes.

\subsubsection{Branch Cropping}

%

Note that not all the non-leaf nodes in the selected branches are closely related to the input caption. We remove non-leaf nodes that are with semantic relevance less than $\Delta$, which is a hyperparameter. Their descendant non-leaf nodes are also removed. To reserve enough frame nodes for training, we do not remove the descendant leaf nodes. Instead, we down-weight them with $\lambda=0.5$. For other leaf nodes, $\lambda=1$.
The remaining leaf nodes and non-leaf nodes represent the critical local information and the global information, respectively.
We multiply the feature of node $i$ and the node's semantic relevance $r_{tv}^i$:
\begin{equation}
  f_{v_{new}}^i=f_{v}^i*r_{tv}^i*\lambda,
  \label{eq:feature_reweight}
\end{equation} 
where $f_{v_{new}}^i$ is the feature vector input into the transformer.
As such, Equation \ref{eq:feature_reweight} considers both local relevance $r_{tv}$ and global relevance $\lambda$ with the language query.

\subsubsection{Self-supervised Tasks}
%
We leverage a transformer encoder for these extracted information and the language query. As shown in the Figure~\ref{fig:schema}, we design two self-supervised tasks as: 1) predicting the masked text features, and masked local/global video information; 2) judging whether the text and the video match.
For the transformer, the input tokens $F_{in}$ consist of the local information, the global information and the text features, which are three types of tokens. We further introduce 2-D position embedding for video tokens and type embedding for all tokens, which are added to the tokens' features.

Then, the features $F_{in}$ are input into the transformer encoder $E$. After encoding, the fusion features $F_{out}$ are output:
\begin{equation}
  F_{out}=E(F_{in}).
  \label{eq:trans_encoder}
\end{equation} 

We predict the original features for masked language tokens and masked video tokens (leaf/non-leaf nodes in the selected branch) using multilayer perceptrons.
\begin{equation}
  {\hat f}_{in}^i=MLP_t(f_{out}^i), \ \ {\hat f}_{in}^j =  MLP_v (f_{out}^j), \label{eq:MLP_video} 
\end{equation} 
where the $MLP_t$ and $MLP_v$ are the multilayer perceptrons for text and video features, respectively. We view masked token modeling as feature regression and adopt L2 distance as the loss function. In addition, there will be a mismatched language query at the rate of 50\%. We propose to predict whether the video and language are matched, \ie, whether the video contains the event described by the language query, based on the output representation of token \texttt{[CLS]}. When the video and the language are not matched, we will not train the model with the one-shot label.

\section{Experiments}
\subsection{Experimental Setup}

\begin{table*}
  \centering
  {\setlength{\tabcolsep}{1.0em}\begin{tabular}{c|cccc|cccc}\hline
  \multirow{2}{*}{Method} & \multicolumn{4}{c|}{Declarative Sentence Grounding} &\multicolumn{4}{c}{Interrogative Sentence Grounding}\\
          & 0.4   & 0.5   & 0.6   & Avg   & 0.4   & 0.5   & 0.6   & Avg  \\\hline
  GroundeR & 24.6 & 18.2 & 13.7 & 18.9    & 25.3 & 18.9 & 14.4 & 19.5     \\
  STPR     & 25.7 & 20.1 & 14.6 & 19.9 & 27.1 & 21.0 & 16.0    & 21.4 \\
  STGRN    & 27.6  & 20.9  & 16.3 & 21.5    & 28.5 & 21.9 & 17.2 & 22.5    \\
  VOGnet   & 32.1 & 24.4 & 19.9  & 25.8    & 33.1 & 25.5 & 20.9 & 26.7 \\
  OMRN   & 34.4 & 27.6 & 21.9  & 28.0    & 35.7 & 28.7 & 23.0 & 29.1    \\\hline
  VOGnet*   & 36.4 & 29.4 & 22.0  & 29.3    & 37.0 & 28.4 & 22.6 & 29.3 \\
  OMRN*   & 39.5 & 30.0 & 22.3  & 30.6    & 38.9 & 30.5 & 24.1 & 31.2 \\ \hline
  
  \textbf{IT-OS}   & \textbf{46.8} & \textbf{35.8} & \textbf{23.2} & \textbf{35.3} & \textbf{46.2} & \textbf{34.6} & \textbf{25.2} & \textbf{35.3}\\\hline
  \end{tabular}}
  \caption{\label{sota_VidSTG}
  Compared with baselines on VidSTVG. It is worth noting that all methods are trained using the one-shot learning. The $*$ represents the baselines use the MDETR as the object detector backbone, which is the same as the IT-OS.}
\end{table*}

\begin{table}
  \centering
  {\setlength{\tabcolsep}{0.7em}\begin{tabular}{c|cccc}\hline
  Method & 0.4 & 0.5 & 0.6 & Avg \\\hline
  GroundeR & 32.1 & 27.8 & 24.3 & 28.1    \\
  STPR     & 33.4 & 28.9 & 25.4 & 29.2 \\
  STGRN    & 35.5  & 30.4  & 26.3 & 30.7 \\
  VOGnet   & 38.8 & 32.7 & 26.9 & 32.8 \\
  OMRN   & 40.1 & 34.5 & 28.4 & 34.4\\\hline
  VOGnet*   & 41.2 & 35.8 & 29.5 & 35.5 \\
  OMRN*   & 45.5 & 37.7 & 30.4 & 37.9\\\hline

  \textbf{IT-OS}   & \textbf{51.9} & \textbf{42.9} & \textbf{33.6} & \textbf{42.8}\\\hline
  \end{tabular}}
  \caption{\label{sota_VID-sentence}
        Compared with baselines on VID-sentence. All methods are trained using one-shot learning. The $*$ represents the MDETR is applied to these baselines as the object detector backbone.}
\end{table}

\vpara{Datasets}
We consider two video grounding benchmarks for evaluation:
(1) {\textit{VidSTG \cite{zhang2020does}}} is a large-scale benchmark dataset for video grounding, which is constructed based on VidOR \cite{shang2019annotating} dataset. VidSTG contains $10,000$ videos and $99,943$ sentences with 
$55,135$ interrogative sentences and $44,808$ declarative  sentences. These sentences describe $79$ types of objects appearing in the videos.
We follow the official dataset split of \cite{zhang2020does}. (2) \textit{VID-sentence \cite{chen2019weakly}} is another widely used video grounding benchmark constructed based on the VID \cite{ILSVRC15} dataset. There are 30 categories and $7,654$ video clips in this dataset. We report the results of all methods on the validation set for the VID-sentence dataset. We obtain similar observations and conclusions on the test set.

\vpara{Implementation Detail}
For video preprocessing, we random resize the frames, and set the max size is $640*640$. The other data augmentation methods, such as random horizontal flip and random size cropping are used at the same time. 
During training, the learning rate is by default $0.00005$, and decays by a factor of $10$ for every $35$ epochs. The batch size is $1$ and the maximum training epoch is $100$. We implement IT-OS in Pytorch and train it on a Linux server.
For model hyperparameters, we set $\lambda=60\%$, and $\Delta=0.7$. Most of the natural language spatial video grounding models use the pretrained detection model as the backbone. Thus, like them,
we choose the official pretrained MDETR \cite{kamath2021mdetr} as the parameter basis for target detection of our IT-OS.

\vpara{Evaluation Metrics}
We follow the evaluation protocol of \cite{chen2019weakly}.  Specifically, we compute the \underline{I}ntersection \underline{o}ver \underline{U}nion (IoU) metric for the predicted spatial bounding box and the ground-truth per frame. The prediction for a video is considered as "accurate" if the average IoU of all frames exceeds a threshold $\alpha$.
The $\alpha$ is set to $0.4$, $0.5$, and $0.6$ during testing.  

\vpara{Baselines}
Since existing video grounding methods are not directly applicable to the one-shot setting, we extend several state-of-the-arts as the baselines. Specifically, to have a comprehensive comparison, we consider 1)fully supervised models, including \textbf{VOGnet} \cite{sadhu2020video}, \textbf{OMRN} \cite{zhang2020object} and \textbf{STGRN} \cite{zhang2020does}; and 2) other widely known methods, including video person grounding \textbf{STPR} \cite{yamaguchi2017spatio}, and visual grounding method, \textbf{GroundeR} \cite{rohrbach2016grounding}.


\subsection{Performance Comparison}

%

The experimental results for one-shot video grounding on VidSTVG and VID-sentence datasets are shown in Table \ref{sota_VidSTG} and \ref{sota_VID-sentence}, respectively. According to the results, we have the following observations:

\begin{itemize}[leftmargin=*]
\item Not surprisingly, although extended to the video grounding setting, baselines that belong to other domains, including video person grounding STPR and visual grounding GroundeR, achieve inferior results on video grounding benchmarks. They lack domain-specific knowledge and might fail to effectively model the spatial-temporal relationships of videos and language queries.
\item IT-OS consistently achieves the best performance on two benchmarks and multiple experimental settings with a large margin improvement. Remarkably, IT-OS boosts the performance (Avg) of the previous state-of-the-art OMRN from nearly $28.0/29.1/34.4$ to $35.3/35.3/42.8$ on VidSTVG and VID-sentence, respectively. It demonstrates the superiority of IT-OS on one-shot video grounding.
\item The baselines are implemented with the backbones used in their original papers, which are different from ours. To further disentangle the sources of performance improvement, we re-implement the best-performing baselines (VOGnet*, and OMRN*) with the same object detection backbone, MDETR, as IT-OS.
Although there is performance improvement with the new backbone, the best-performing baseline OMRN*, still underperforms IT-OS by over $4$ points for the average accuracy on all datasets.
It further reveals the effectiveness of our novel model designs eliminating interference with different pre-training parameters. We attribute the improvement to the end-to-end modeling, where different modules can simultaneously benefit from each other. In addition, the proposed information tree alleviates the negative effects of irrelevant frames, and effectively models the interactions between the video global/local information and the language query. Several self-supervised learning tasks based on the information tree enhance the representation learning under limited one-shot labels.
%
\end{itemize}

\begin{table}[t]
  \centering
  {\setlength{\tabcolsep}{0.6em}\begin{tabular}{c|cccc}\hline
  Method & 0.4 & 0.5 & 0.6 & Avg\\ \hline
  GroundeR & 42.72 & 33.77 & 27.05 & 34.51 \\
  STPR     & 47.95 & 36.19 & 30.41 & 38.18 \\
  STGRN    & 49.25  & 44.03 & 34.89 & 42.72 \\
  VOGnet     & 53.17 & 43.47 & 33.77 & 43.47 \\
  OMRN   & 55.22 & 46.64 & 37.50 & 46.45 \\\hline
  IT-OS (OS)   & 51.87 & 42.91 & 33.58 & 42.79  \\\hline
  \end{tabular}}
  \caption{\label{sota_VID-sentence_supervised}
  Compared with the baselines on VID-sentence. The baselines are trained using fully supervised learning. The OS represents the IT-OS trained under the one-shot settings.}
\end{table}

\begin{table*}[t]
  \centering
  {\setlength{\tabcolsep}{0.8em}\begin{tabular}{ccc|cccc|cccc}\hline
  \multicolumn{3}{c|}{\multirow{2}{*}{}} & \multicolumn{4}{c|}{Declarative Sentence Grounding} &\multicolumn{4}{c}{Interrogative Sentence Grounding}\\
  $\Gamma_{self}$   &$\Gamma_{tree}$ & $\Gamma_{crop}$     & 0.4   & 0.5   & 0.6   & Avg   & 0.4   & 0.5   & 0.6   & Avg  \\\hline
          &           &         & 39.00           & 30.52          & 17.61          & 29.05   & 38.78           & 28.75          & 19.67          & 29.07     \\
  \checkmark &        &      & 40.52           & 32.32          & 18.83          & 30.56    & 40.82          & 31.44          & 20.66          & 30.97   \\
          &\checkmark &    & 42.34           & 32.65          & 20.35          & 31.78    & 42.26  & 32.02 & 21.89          & 32.06    \\
  \checkmark&  \checkmark  &     & 44.16           & 33.38          & 21.11          & 32.89   & 44.55          & 33.78          & 23.19          & 33.84   \\
  &\checkmark&\checkmark& 44.77           & 34.62          & 22.93          & 34.11    & 44.30          & 33.23          & 24.17          & 33.90  \\
  \checkmark&\checkmark&\checkmark & \textbf{46.75} & \textbf{35.81} & \textbf{23.23} & \textbf{35.26} & \textbf{46.16} & \textbf{34.55} & \textbf{25.19} & \textbf{35.30}\\\hline
  \end{tabular}}
  
  \caption{\label{ablationstudy}
  Ablation study on VidSTG dataset.}
  \end{table*}

  {\setlength{\tabcolsep}{0.3em}\begin{table}[t]
  \centering
  \begin{tabular}{ccc|cccc}\hline
  $\Gamma_{self}$   &$\Gamma_{tree}$ & $\Gamma_{crop}$    & 0.4   & 0.5   & 0.6   & Avg  \\\hline
          &           &         & 44.40           & 35.07          & 27.24          & 35.57   \\
  \checkmark &        &      & 46.64           & 36.38          & 28.54          & 37.19  \\
          &\checkmark &    & 47.95           & 38.99          & 29.85          & 38.93   \\
  \checkmark&  \checkmark  &    & 49.44           & 40.30          & 31.16          & 40.30   \\
  & \checkmark&\checkmark& 50.19           & 40.49          & 32.46          & 41.04   \\
  \checkmark&\checkmark&\checkmark & \textbf{51.87} & \textbf{42.91} & \textbf{33.58} & \textbf{42.79} \\\hline
  \end{tabular}
  
  \caption{\label{ablationstudyVIDS}
  Ablation study on VID-sentence dataset.}
  \end{table}
}

\subsection{Comparison with Fully Supervised Methods}
We are interested in 1) how different baselines perform under fully supervised settings; 2) how one-shot IT-OS perform compared to these baselines. Towards this end, we train multiple baselines and IT-OS with all labels on the VID-sentence dataset.
 The experiment results are shown in Table~\ref{sota_VID-sentence_supervised}. From the table, we have the following findings:
\begin{itemize}[leftmargin=*]
  \item Remarkably, the performance gap between one-shot IT-OS and the fully supervised OMRN is less than $4\%$. Such a minor gap demonstrates the effectiveness of IT-OS on learning with limited annotations. This is significant and practical merit since we are more likely to have a limited annotation budget in real-world applications.  
\item Surprisingly, one-shot IT-OS can still outperform some weak baselines such as GroundeR and STPR. These results reveal the necessity of end-to-end modeling for video grounding.
\end{itemize}

\subsection{Ablation Study}
We are interested in how different building blocks contribute to the effectiveness of IT-OS. To this end, we surgically remove several components from IT-OS and construct different architectures. The investigated components include information tree ($\Gamma_{tree}$), the branch cropping ($\Gamma_{crop}$), and the self-supervised training ($\Gamma_{self}$). 
It is worth noting that the other components cannot be deleted independently except the branch cropping. Thus, we don't conduct an ablation study for them. 
Results on VidSTG and VID-sentence datasets are shown in Table~\ref{ablationstudy} and Table~\ref{ablationstudyVIDS}, respectively. There are several observations: 
\begin{itemize}[leftmargin=*]
	\item Overall, removing any component incurs a performance drop, demonstrating the necessity and effectiveness of the information tree, branch search \& cropping, and self-supervised training.
	\item Stacking multiple components outperform the architecture with a single component. This result reveals that the proposed components can benefit from each other in end-to-end training and jointly boost one-shot video grounding. 
\end{itemize}


\subsection{Case Study}

We conduct a case study to visually reveal the ability of the IT-OS in detail. Specifically, we random sample $3$ videos from the datasets, and sample $6$ frames from each video to visualize.

\begin{figure}[t]
  \centering
   \includegraphics[width=1.0\linewidth]{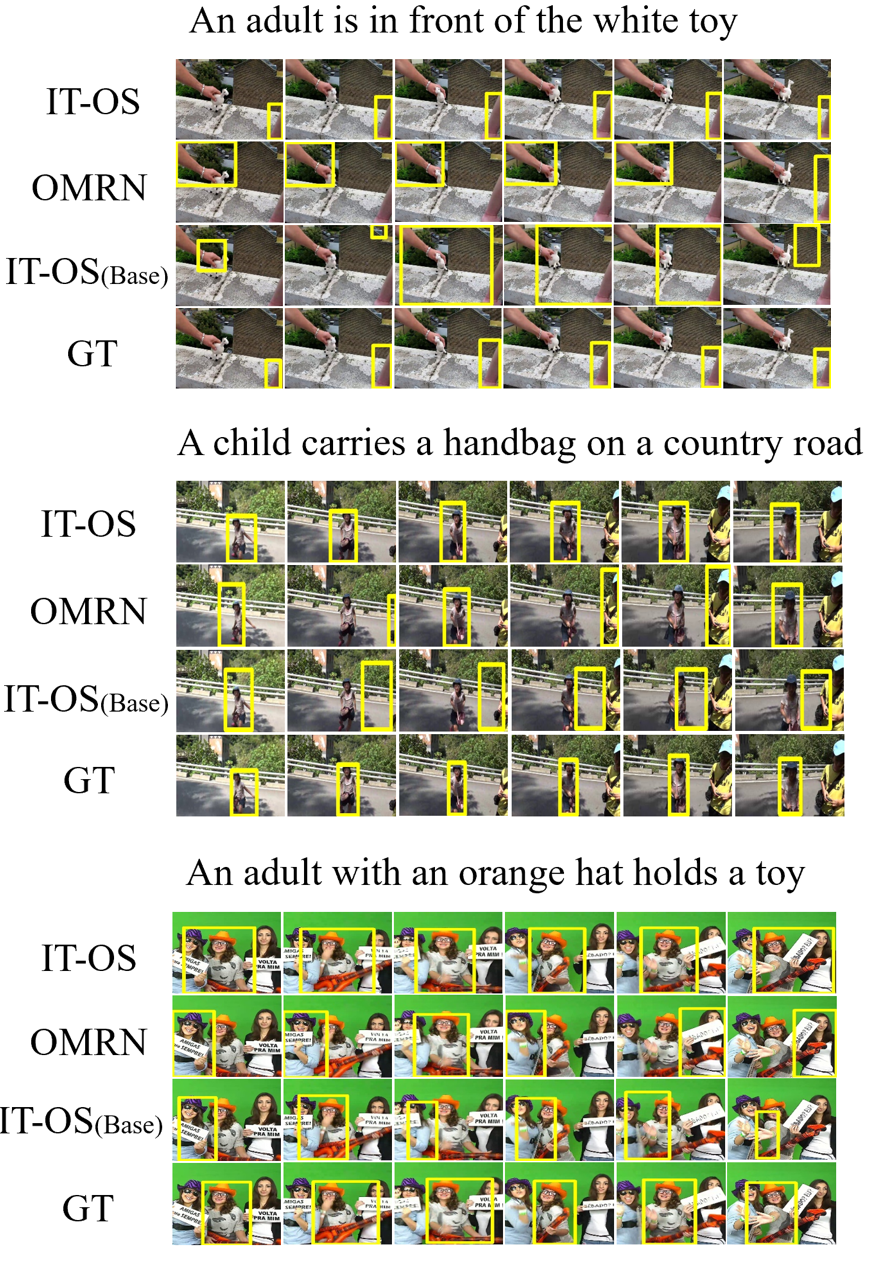}

   \caption{Examples of the detection result visualization. The IT-OS(Base) represents the IT-OS model without the self supervised module and the informaiton tree. The GT represents the target labels.}
   \label{fig:4_1}
\end{figure}

We compare our IT-OS model with the baseline method, OMRN, and the fundamental ablation model of the IT-OS, which is removed from the self-supervised module and the information tree. As shown in Figure~\ref{fig:4_1}, we have the following key findings:
(1) The IT-OS detects the more accurate one from all objects of the video than the best performing previous method. It demonstrates the better representation extraction and analysis capabilities of our model.
(2) Even if the target object is selected correctly, the IT-OS localizes a more precise spatial area compared with the previous two stages method. The results reflect the end-to-end model, IT-OS, has more accurate domain knowledge through training the whole model on the target dataset.
(3) After adding the information tree and the self-supervised module, the IT-OS outputs more precise bounding boxes. It reveals that combining the two modules introduce stronger supervision signals for model training so that the model has stronger detection ability.

\section{Conclusion}

In this paper, we introduce the one-shot learning into the natural language spatial video grounding task to reduce the labeling cost. To achieve the goal, the main point is to make full use of only one frame label for each video. The invalid frames unrelated to the input text and target objects bring confounding to the one-shot training process. We design an end-to-end model (IT-OS) via the information tree to avoid it. Specifically, the information tree module merges frames with similar semantics into one node. Then, by searching the tree and cropping the invalid nodes, we can get the complete and valid semantic unit of the video. Finally, two self-supervised tasks are used to make up the insufficient supervision. 

\section*{Acknowledgements}
This work is supported in part by the National Natural Science Foundation of China (Grant No.62037001, No.61836002, No.62072397). This work is also partially funded by Hangzhou Hikvision Digital Technology.

\bibliography{anthology,custom}
\bibliographystyle{acl_natbib}




\end{document}